\documentclass[twocolumn]{article}
\usepackage{spconf,amsmath, amssymb, graphicx,hyperref,caption, booktabs, algorithm, algpseudocode, multirow,bbold}

\usepackage{listings}
\usepackage{xcolor}
\definecolor{lightgray}{gray}{0.95}
\lstdefinestyle{grayblock}{
  backgroundcolor=\color{lightgray},
  basicstyle=\ttfamily\small,
  frame=single,
  breaklines=true,
  postbreak=\mbox{\textcolor{red}{$\hookrightarrow$}\space},
  tabsize=2,
  showstringspaces=false
}

\title{STENCIL: Subject-Driven Generation with Context Guidance}

\name{Gordon Chen\textsuperscript{1,2}, Ziqi Huang\textsuperscript{1}, Cheston Tan\textsuperscript{2}, Ziwei Liu\textsuperscript{1}}

\address{\textsuperscript{1}S-Lab, Nanyang Technological University, \textsuperscript{2}CFAR, IHPC, A*STAR\\[0.5em]
        \href{https://gordonchen19.github.io/STENCIL.github.io/}{\texttt{stencil.github.io}}
}

\usepackage[dvipsnames]{xcolor}

\newcommand{\ziwei}[1]{{\color{ForestGreen}\textbf{(ziwei)}}}

\begin{document}

\twocolumn[{%
            \renewcommand\twocolumn[1][]{#1}%
            \maketitle
            \vspace{-2em}
            \begin{center}
                \centering
                \includegraphics[width=0.99\textwidth]{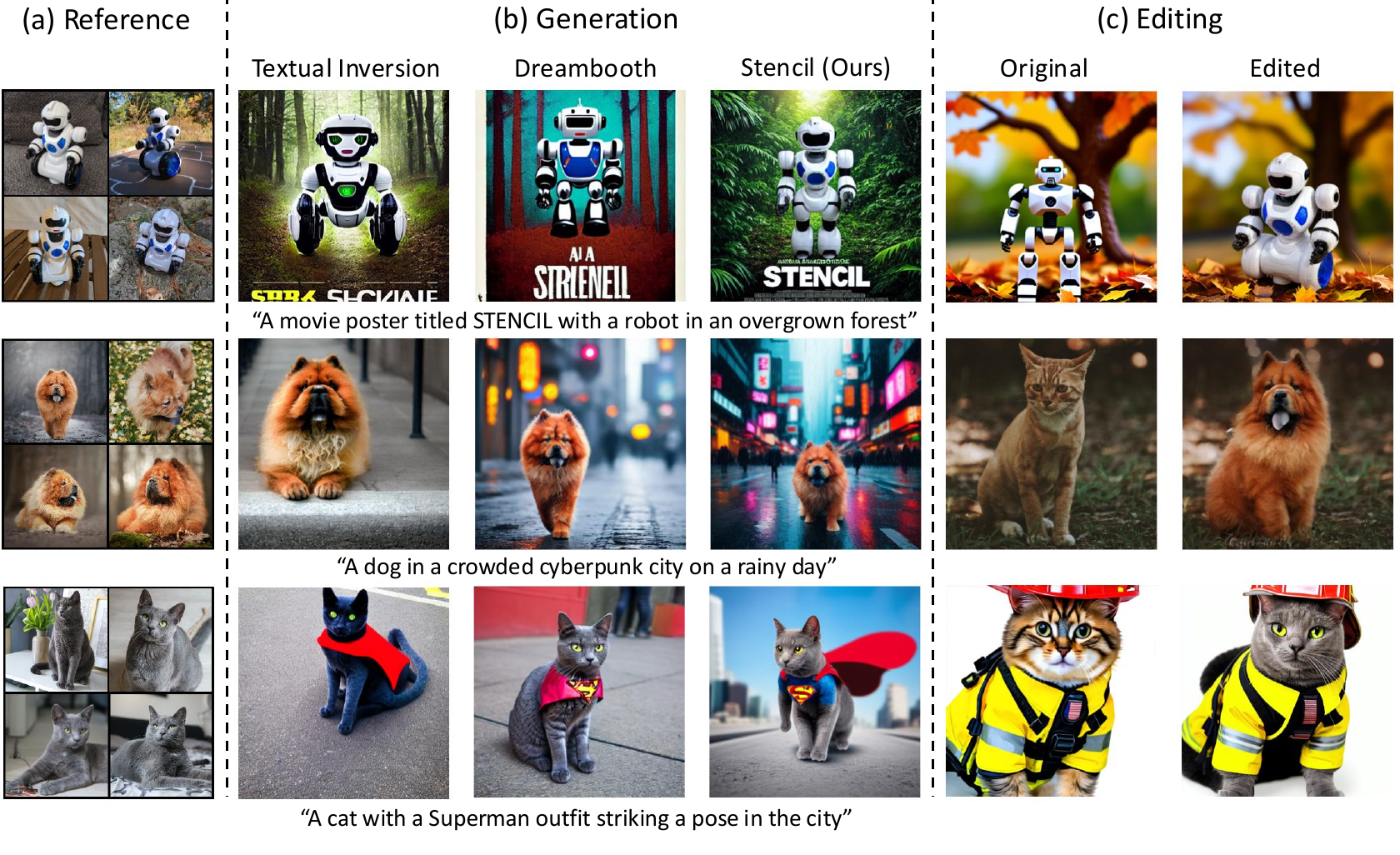}
                \vspace{0.1cm}
                \vskip -0.2cm
                \captionof{figure}{\textbf{Overview of \textit{Stencil}}. Given a few (a) reference images,  Stencil achieves (b) subject-driven generation and (c) subject editing with high textual and subject fidelity in just 100 fine-tuning steps. 
                }
                \label{teaser}
            \end{center}%
        }]

\begin{abstract}
Recent text-to-image diffusion models can generate striking visuals from text prompts, but they often fail to maintain subject consistency across generations and contexts. One major limitation of current fine-tuning approaches is the inherent trade-off between quality and efficiency. Fine-tuning large models improves fidelity but is computationally expensive, while fine-tuning lightweight models improves efficiency but compromises image fidelity. Moreover, fine-tuning pre-trained models on a small set of images of the subject can damage the existing priors, resulting in suboptimal results. To this end, we present Stencil, a novel framework that jointly employs two diffusion models during inference. Stencil efficiently fine-tunes a lightweight model on images of the subject, while a large frozen pre-trained model provides contextual guidance during inference, injecting rich priors to enhance generation with minimal overhead. Stencil excels at generating high-fidelity, novel renditions of the subject in less than a minute, delivering state-of-the-art performance and setting a new benchmark in subject-driven generation.

\end{abstract}


\begin{figure*}[htb]
            \begin{center}
                \includegraphics[width=0.99\textwidth]{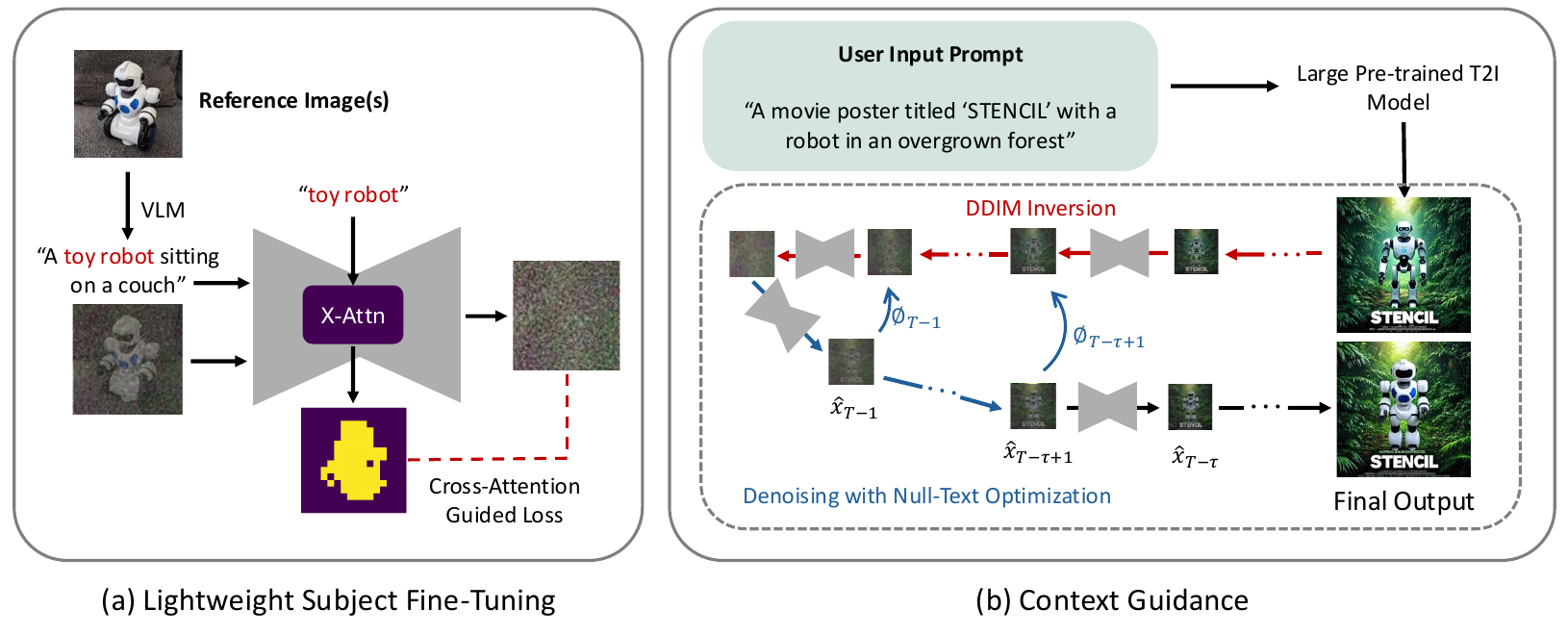}
                \captionof{figure}{\textbf{\textit{Stencil} Framework.} 
                \textbf{(a) Lightweight Subject Fine-Tuning.}  
                We fine-tune a lightweight text-to-image diffusion model on the reference image(s) of the subject.  
                The Cross-Attention Guided Loss is applied so that gradients are computed only in regions influenced by the subject token(s) (e.g.\ “\textit{toy robot}”).  
                \textbf{(b) Context Guidance.}  
                At inference time, we use a large pre-trained text-to-image model to generate an image corresponding to the user prompt. The image is inverted and denoised with null-text optimisation by the lightweight fine-tuned model, producing the final image that preserves the rich contextual features from the large pre-trained model and adapts the subject appearance to align with the references via the small fine-tuned model.}
                \label{pipeline}
            \end{center}
\end{figure*}

\section{Introduction}
\label{sec:intro}

Text-to-image (T2I) diffusion models \cite{saharia2022photorealistic} have achieved remarkable success in producing high-quality, text-aligned images. Recently, subject-driven generation has emerged as a pivotal research area, enabling users to personalize subjects by providing reference images as input to T2I models. A widely adopted strategy is to fine-tune T2I models on the reference images. However, these methods face notable limitations. Fine-tuning a lightweight diffusion model is efficient but degrades image quality. In contrast, fine-tuning a larger diffusion model \cite{esser2024scaling, podell2023sdxl} yields higher-fidelity images, but is more computationally expensive. Moreover, fine-tuning pre-trained models on a few images can easily damage its existing priors. This underscores the need for more robust fine-tuning approaches. To this end, we propose Stencil, a novel framework that jointly employs two diffusion models during inference. Stencil leverages a lightweight model to efficiently learn new subject priors and a large pre-trained model to inject rich contextual priors to enhance generation. This collaboration between two models enables efficient fine-tuning and high-fidelity outputs that neither model could achieve alone. Furthermore, we introduce the Cross-Attention Guided Loss during fine-tuning, which computes loss only on subject-relevant regions of the reference images, reducing optimization complexity and enabling faster convergence. Stencil achieves state-of-the-art (SOTA) results in generation while being the most cost-effective framework.  Our main contributions are as follows:

{\setlength{\parskip}{0pt}
• We propose Context Guidance, a novel technique that employs two diffusion models jointly during inference, achieving superior results over using a single model.

•  We propose the Cross-Attention Guided Loss function, where we only optimize subject-relevant areas of the reference images to improve fine-tuning stability and efficiency.

• Our extensive experiments have validated the robustness of our approach, achieving SOTA results. 
}

\section{Related Works}
\label{sec:related works}

Personalizing diffusion models for subject-driven generation broadly falls into two families: (i) test-time finetuning of the model on a few reference images \cite{ruiz2023DreamBooth, gal2022image, kumari2023multi,avrahami2023break, huang2024reversion, tewel2023key, hua2023dreamtuner}, and (ii) tuning-free encoders that inject subject information without updating the base model \cite{li2024blip,zhang2024ssr,wei2023elite,patel2024lambda, ye2023ip, shi2024instantbooth, wu2024difflora, arar2023domain, gal2024lcm}.

\subsection{Test-time fine-tuning}

Textual Inversion \cite{gal2022image} optimizes a new token embeddings to bind a subject representation to a prompt slot, avoiding full model updates but struggling with composition and strong identity fidelity. Perfusion \cite{tewel2023key} introduced key-locked concept vectors into the cross-attention, enabling efficient subject injection at the cost of subject fidelity. DreamBooth \cite{ruiz2023DreamBooth} fine-tunes the denoising U-Net to bind the subject representation to a class token, achieving high fidelity outputs at a notable computational cost and potential prior drift. Custom Diffusion \cite{kumari2023multi} improves efficiency by limiting fine-tuning to the cross-attention layers of the U-Net. DreamTuner \cite{hua2023dreamtuner} fine-tunes lightweight LoRA adapters in the cross-attention layers to capture a subject's identity. While U-Net fine-tuning tends to yield the strongest subject fidelity, it scales poorly to large backbones due to computational costs and can erode pre-trained priors when data is scarce. 

\subsection{Tuning-free adapters}
Other approaches inject subject features into the cross-attention layers of a frozen diffusion model. DiffLoRA \cite{wu2024difflora} trains a hypernetwork to predict subject-specific LoRA weights directly from reference images. InstantBooth \cite{shi2024instantbooth} encodes reference images into global subject embeddings, while BLIP-Diffusion \cite{li2024blip}  extends this idea with a multi-modal encoder that jointly models image and text for subject representation. SSR-Encoder \cite{zhang2024ssr} uses a token-to-patch aligner to encode relevant regions to prevent the background features from leaking into the subject representation. These approaches remove the need for test-time fine-tuning, but typically require large-scale training data and risk memorization or leakage from training subjects, limiting generalization.

\section{Preliminaries} 
\label{sec:preliminaries}

\subsection{Text-to-Image Latent Diffusion Models.} Diffusion models are a class of generative models that progressively transform pure Gaussian noise $x_T$ into a target image $x_0$ through iterative denoising steps. Each model consists of a denoising network $f_\theta(x_t, t, \psi(P))$, traditionally a U-Net, conditioned on the text embedding $\psi(P)$, to predict the noise residual $\epsilon_t$ of $x_t$ at time-step $t$, enabling the reconstruction of a slightly de-noised sample $x_{t-1}$. Latent diffusion models \cite{rombach2022high} reduce compute complexity by applying the diffusion process on a lower-dimensional latent space $z_t$.  The overall loss function for training such a denoising network is computed as:
\begin{equation}
L = \mathbb{E}_{z_0, \epsilon , t, \psi(P)} \left[ \| \epsilon - f_\theta(z_t, t, \psi(P)) \|_2^2 \right]
\end{equation}

\subsection{Cross-Attention Mechanism.} In cross-attention, the deep spatial features $\phi(z_t)$ are linearly projected to a query $Q=\ell_Q\phi(z_t)$, key $K=\ell_K\psi(P)$, and value $V=\ell_V\psi(P)$ matrix via learned projections $\ell_Q,\ell_K,\ell_V$ respectively. The attention map is formulated as:

\begin{equation}
M = \text{Softmax}\left(\frac{QK^\top}{\sqrt{d}}\right)
\end{equation}\\[0.1em]
where $d$ is the latent projection dimension of the keys and queries. The entry $M_{ij}$ defines the weight of the $j$-th token on the pixel $i$. Intuitively, cross-attention maps bind each text token to specific regions of the image, which guides the placement of textual elements in the generated image \cite{hertz2022prompt}. The attention output $\widehat{\phi}(z_t) = MV$, updates $\phi(z_t)$ and is propagated to the subsequent layers of the U-Net. \\[0.1em]

\subsection{Inversion with Null-Text Optimization.} Inversion is a core technique allowing us to retrieve the noise vector corresponding to a given image, such that the forward diffusion process on the noise vector reproduces the given image. However, due to the stochasticity of diffusion, the reconstructed image may look different from the original image. Null-text optimization \cite{mokady2023null} solves this problem by leveraging DDIM inversion to establish a pivot trajectory. At each denoising time-step, the unconditional embeddings $\emptyset_t$ are optimized to minimize the deviations from the pivot trajectory, enabling perfect reconstructions of the original image. The objective is:

\begin{equation}
\min_{\emptyset_{t}} \left\|z_{t-1}^{*} - \text{DDIM}(\bar{z}_t, \hat{\epsilon}_t, t) \right\|_2^2, \quad \hat{\epsilon}_t = f_\theta(\bar{z}_t, t, \emptyset_t, \psi(P))
\end{equation}

\begin{figure*}[htb]
            \begin{center}
                \centering
                \includegraphics[width=0.99\textwidth]{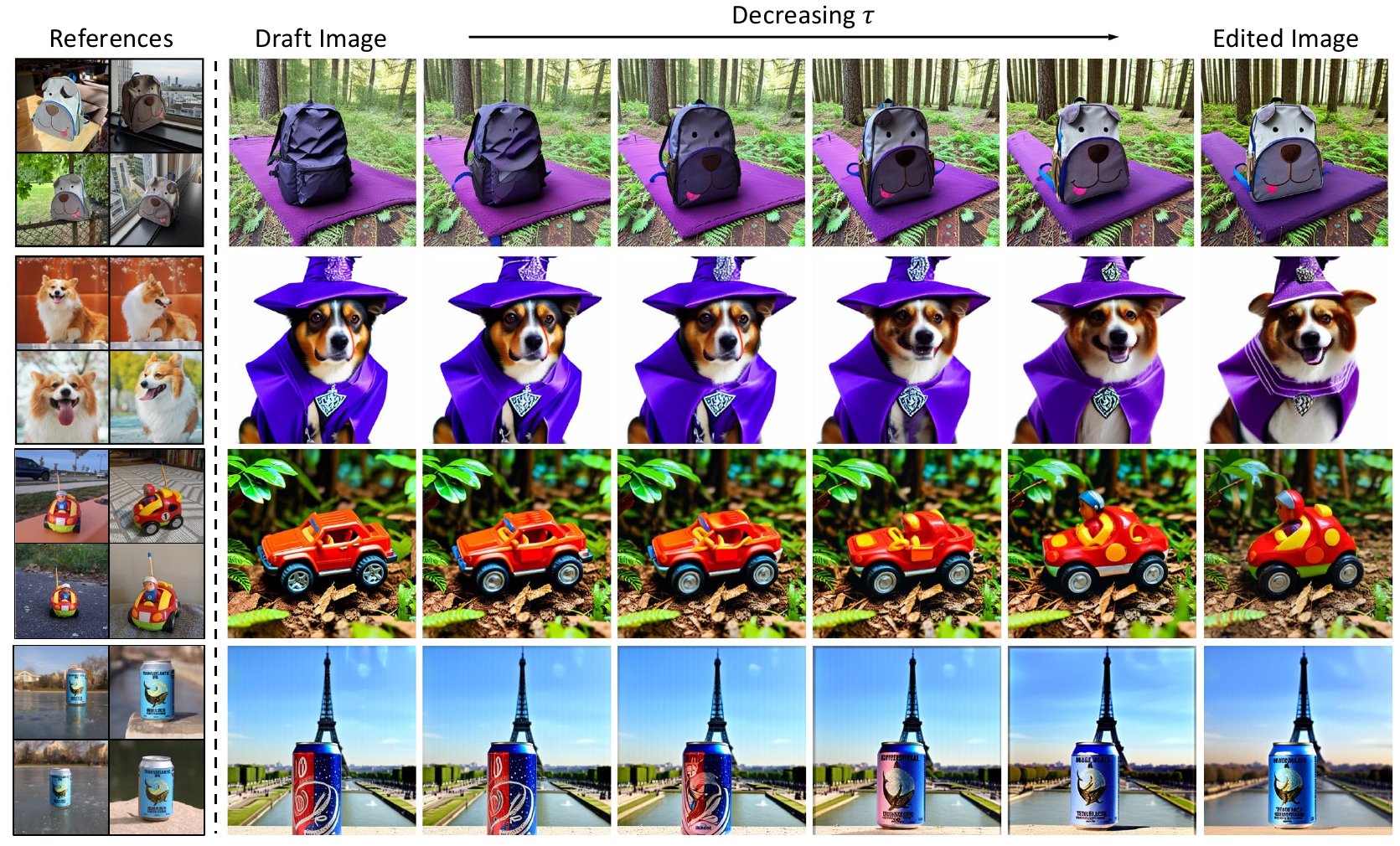}
                \captionof{figure}{\textbf{Modifying $\tau$.} Decreasing  $\tau$ will allow the fine-tuned model to impose its learned subject priors over more denoising steps while still preserving the contextual priors of the original image. An appropriate choice of $\tau$ balances subject fidelity against background preservation. A small value favors stronger subject adaptation at the risk of altering context, whereas a large value preserves the original image but limits the subject modification.}
                \label{null-text}
            \end{center}
\end{figure*}

\section{STENCIL}

Fig. \ref{pipeline} illustrates Stencil's method framework, consisting of an initial fine-tuning stage with the Cross-Attention Guided Loss (Sec. \ref{sec:Masked Loss} ) followed by Context Guidance at inference time (Sec. \ref{sec:Context Guidance}). Additional implementation details are discussed in \textit{Supplementary Materials}.

\subsection{Cross-Attention Guided Loss Function}
\label{sec:Masked Loss}
The traditional Mean-Squared Error (MSE) loss treats subject and background pixels in the reference images indiscriminately. As a result, applying the MSE loss across the entire reference image can complicate optimization and allow background features to leak into the learned representation of the subject. This may also cause spurious background artifacts in the final generation. The Cross-Attention Guided Loss effectively addresses this problem by restricting optimization to relevant pixels in the reference images, thereby simplifying training and preserving a clean subject representation.

Specifically, we use a vision-language model (VLM) to generate a caption $C_i$ for the reference image $I_i$. We then add $t$ time-step noise to the encoded image and feed the resulting noisy $z_t$ latent into the frozen U-Net, conditioned on $C_i$. During this forward pass, we save the cross-attention maps corresponding to the subject token $S$ from all heads and layers, up-sample them to the same latent resolution, compute the mean and normalize the values to obtain an aggregated cross-attention map, $\widehat{M}_i^S$. In this map, higher values indicate spatial regions where $S$ receives stronger attention, while lower values correspond to background or irrelevant areas that contribute little to the subject representation. We fine-tune the U-Net on ($I_i$, $C_i$, $\widehat{M}_i^S$) using the proposed Cross-Attention Guided Loss, defined as:
\begin{equation}
L = \mathbb{E}_{z_0, \epsilon, t, \psi(C)} \left[ \| \mathbb{1}_{\widehat{M}_S > p_{t}} \cdot(\epsilon - f_\theta(z_t, t, \psi(C)) ) \|_2^2 \right]
\end{equation}
where  $\mathbb{1}_{\widehat{M}_S > p_{t}}$ denotes a binary mask matching the resolution of the image latent, that marks pixels whose attention weight from $S$ exceeds the threshold $p_t$. Pixels with a value of 1 are treated as subject-relevant, while those with 0 correspond to background. We apply this mask element-wise to the loss, restricting optimization to subject regions and guiding the U-Net to learn a clean subject representation.

\subsection{Context Guidance}
\label{sec:Context Guidance}

Fine-tuning  a pre-trained model on a small dataset can degrade its existing priors, often leading to poor performance afterwards. To address this limitation, we propose Context Guidance, which injects pre-trained priors at inference time through a separate model. Specifically, we jointly leverage a small fine-tuned T2I model $\hat{f}$ (Sec. \ref{sec:Masked Loss}) as well as a large pre-trained T2I model $F$. In this setup, $\hat{f}$ enforces subject consistency, while $F$ enhances the image with its rich pre-trained contextual priors, enabling high-quality subject-driven generation without the heavy computational cost of fine-tuning $F$. 

We first use $F$ to generate an image $I$ conditioned on the user's prompt $\mathcal{P}$.  Our goal is to replace the subject in $I$ with that from the reference images while preserving the rest of the scene as faithfully as possible. To this end, we encode $I$ into its latent space and perform DDIM Inversion with null-text optimization using $\hat{f}$, obtaining the inverted latent $\hat{\mathbf{z}}_T$ and the optimized unconditional embeddings $\emptyset$ at each denoising step. Injecting $\emptyset$ at every denoising step of $\hat{\mathbf{z}}_T$ would simply reconstruct $I$, which is not what we want. Instead, restricting $\emptyset$  injections to the early denoising steps allows the fine-tuned model to impose its learned subject priors in the later stages of denoising. Formally, we inject $\emptyset$ for the first $\tau$  time-steps. We denote this operation as:

\begin{equation}
\epsilon_{t} =
\begin{cases}
    \hat{f}_\theta(\mathbf{\hat{z}}_t, t, \psi(P), \emptyset_t) & \text{if } \tau > T-t \\
    \hat{f}_\theta(\mathbf{\hat{z}}_t, t, \psi(P)) & \text{otherwise}
\end{cases}
\end{equation}

We show empirically in Fig. \ref{null-text} that this operation modifies the subject's appearance in $I$ based on the subject priors learned by $\hat{f}$,  while keeping the rest of the image as faithful to $I$ as possible.  Effectively, Context Guidance allows us to produce images that combine the subject representation priors from $\hat{f}$  with the rich contextual priors from $F$.

\begin{figure*}[htb]
            \begin{center}
                \centering
                \includegraphics[width=0.99\textwidth]{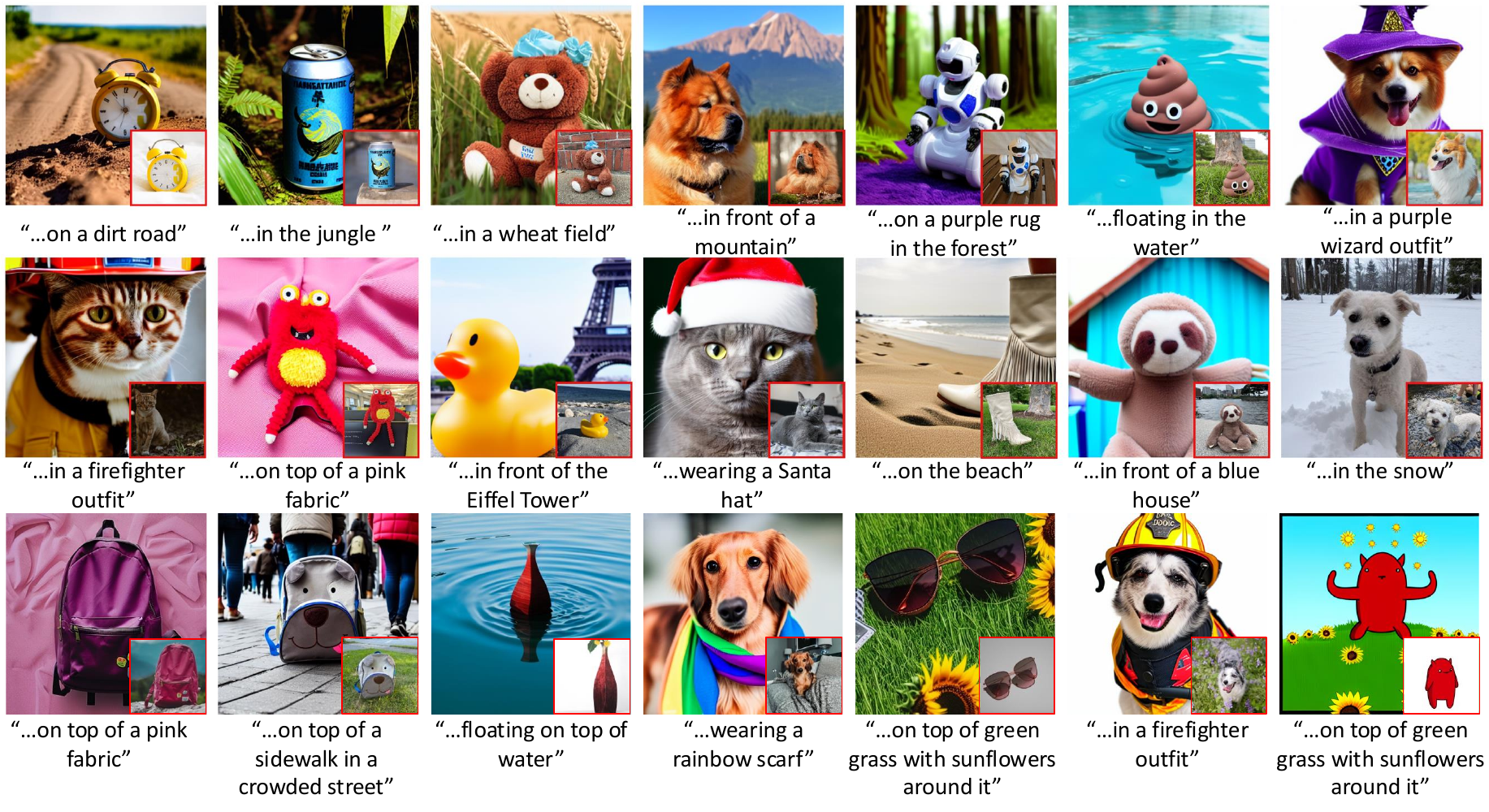}
                \captionof{figure}{\textbf{Qualitative Results}. Stencil produces high-quality images while maintaining subject consistency with the reference (bottom right).
                }
                \label{results}
            \end{center}
\end{figure*}

\begin{table*}
\centering

\caption{
\textbf{Quantitative comparison on DreamBench.} 
\textbf{Bold} entries indicate the top-performing method for each evaluation metric. Experimental results are referenced from the original papers as well as \cite{patel2024lambda}.
}
\vskip -0.2cm
\label{tab:dreambench_results}

\begin{tabular}{c|c|c|cc|c}\toprule
\textbf{Type} & \textbf{Method} & \textbf{Base Model} & \textbf{Subject Consistency ($\uparrow$)} & \textbf{Text Alignment ($\uparrow$)} & \textbf{GPU Hours ($\downarrow$)} \\\midrule
\multirow{7}{*}{Fine-tuning Free} 
& \textbf{ELITE} & SDv1.4 & 0.621 & 0.293 & 336 \\
& \textbf{BLIP-Diffusion} & SDv1.5 & 0.594 & 0.300 & 2304 \\
& \textbf{IP-Adapter} & SDXL & 0.613 & 0.292 & 672 \\
& \textbf{Kosmos-G} & SDv1.5 & 0.618 & 0.250 & 12300 \\
& \textbf{Emu2} & SDXL & 0.563 & 0.273 & - \\
& \textbf{$\lambda$-eclipse} & Kv2.2 & 0.613 & 0.307 & 74 \\
& \textbf{SSR-Encoder} & SDv1.5 & 0.612 & 0.308 & - \\\midrule
\multirow{4}{*}{Fine-tuning} 
& \textbf{Textual Inversion} & SDv1.5 & 0.569 & 0.255 & 1 \\
& \textbf{DreamBooth} & SDv1.5 & 0.668 & 0.305 & 0.2 \\
& \textbf{Custom Diffusion} & SDv1.5 & 0.643 & 0.305 & 0.2 \\
& \textbf{Stencil (Ours)} & SDv1.5 & \textbf{0.671} & \textbf{0.328} & \textbf{0.1} \\
\bottomrule
\end{tabular}

\end{table*}

\section{Experiments}

\noindent\textbf{Experiment Setup.} We use Stable Diffusion V1-5 \cite{rombach2022high} as our base diffusion model. At inference, we use Stable Diffusion 3 Medium \cite{esser2024scaling} to provide Context Guidance. We use GPT-4o \cite{hurst2024gpt} to generate captions for each reference image. Reference images are resized to 512x512 resolution, center-cropped, and normalized. We set the threshold $p_t=0.2$ for our Cross-Attention Guided Loss. Fine-tuning is then performed in batches of 6 on a single A100 GPU for 100 iterations at a learning rate of 2e-5. Inference was performed with DDIM sampling \cite{song2020denoising}, with a step size of 50 and a guidance scale set to 7.5. We set $\tau=60$ for Context Guidance.

\noindent\textbf{Evaluation Metrics.} We evaluate Stencil on the DreamBench dataset \cite{ruiz2023DreamBooth}, consisting of 30 subjects each represented by 4-7 reference images and tested across 25 prompts. We assess the image quality along two key dimensions:  \textit{subject consistency}, which measures how closely the generated subject resembles the true subject using DINO scores (computed as the average pairwise cosine similarity between ViT-S/16 DINO embeddings of the generated and reference images), and \textit{text alignment}, which evaluates how accurately the generated image reflects the user prompt using CLIP-T scores (computed as the average cosine similarity between CLIP embeddings of the prompt and the generated image).

\begin{table}[h]
    \centering
    \begin{tabular}{c|cc}\toprule
        \textbf{Method} & \textbf{Subject Consistency} & \textbf{Text Alignment} \\\midrule
        \textbf{Stencil (Ours)} & 0.782 & 0.764 \\
        \textbf{DreamBooth} & 0.153 & 0.173 \\
        \textbf{Undecided} & 0.064 & 0.062 \\
    \bottomrule
    \end{tabular}
    \caption{\textbf{User Study.} We compare Stencil to DreamBooth. Values indicate user preferences in decimal values.}
    \label{fig:User Study}
\end{table}

\section{Experiment Results}
\label{sec:Experiment Results}

\subsection{Quantitative Evaluation}
\label{sec:Quantitative Evaluation}

Table \ref{tab:dreambench_results} summarizes our quantitative evaluations. Stencil outperforms all previous methods in both text and subject fidelity. To our best knowledge, Stencil is the new SOTA among open-source methods. Stencil noticeably outperforms the others at producing semantically accurate images. This shows that Context Guidance is a robust mechanism for enhancing generation quality at no cost of subject consistency. Stencil is also the most cost-effective framework. Despite the added inference overhead from Context Guidance, it achieves the lowest end-to-end GPU time. This shows that the Cross-Attention Guided Loss can enable faster subject convergence, making it an effective loss for learning clean subject representations.

\subsection{Qualitative Evaluation}
\label{sec:Qualitative Evaluation}
Fig. \ref{results} showcases images generated by Stencil. Compared to other methods, Stencil excels at generating diverse and high-fidelity layouts while maintaining subject fidelity. This shows that Context Guidance can mitigate overfitting to the reference images by leveraging pre-trained priors. Table \ref{fig:User Study} presents results from our user study consisting of 30 participants, where Stencil decisively outperforms DreamBooth in both text alignment and subject consistency.

\begin{figure}[t]
    \centering
    \includegraphics[width=.93\linewidth]{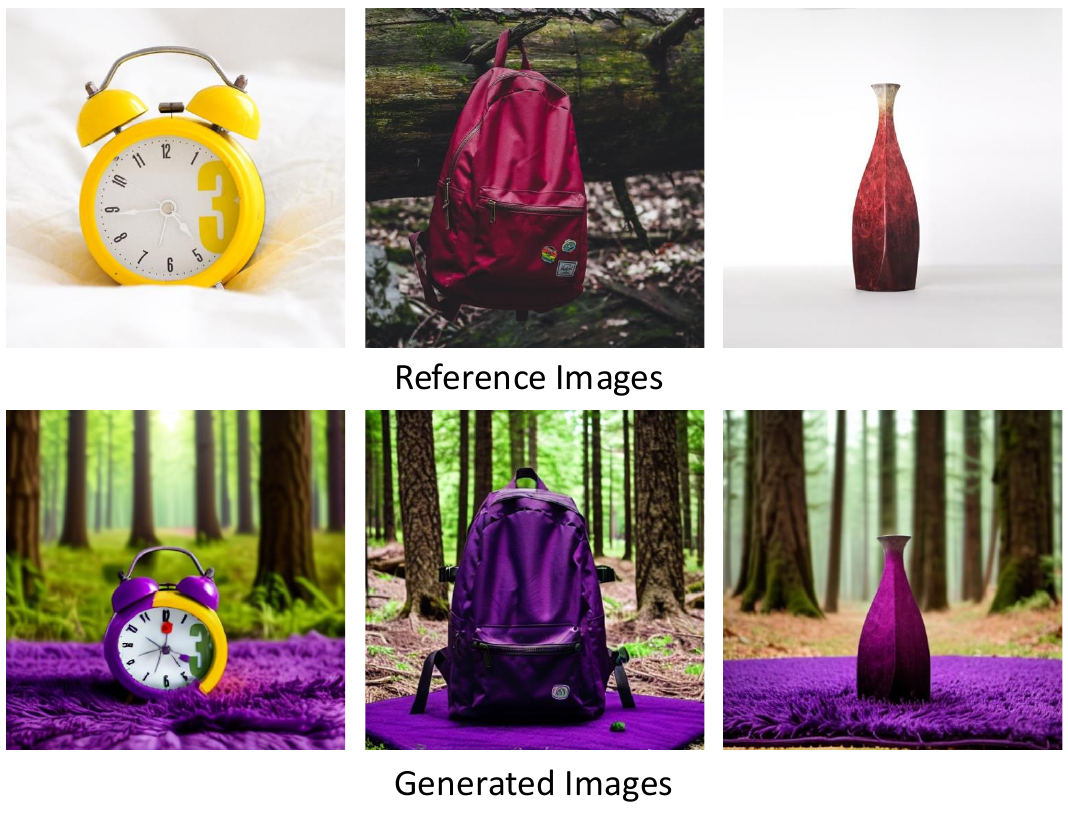}
    \captionof{figure}{\textbf{Failure Case.} Stencil can sometimes inherit the limitations of its base diffusion model.
    }
    \label{fig:failure cases}
\end{figure}

\subsection{Discussions and Limitations}

A detailed discussion of Stencil's applications and limitations are provided in our \textit{Supplementary Materials}. Stencil supports a wide range of applications, including age progression/regression, expression editing, accessorization, perspective-conditioned generation, pose manipulation, and style transfer.

As illustrated in Fig. \ref{fig:failure cases}, Stencil inherits certain limitations from its underlying diffusion models, most notably the tendency for local features to be applied globally. Additionally, some subjects (e.g.,  animals) are easier to learn than others (e.g., human faces), due to differences in training distribution.

\section{Conclusion}

In this paper, we introduced Stencil, an efficient fine-tuning approach for subject-driven generation. Stencil incorporates two key innovations: first, the Cross-Attention Guided Loss function that directs the network’s learning towards subject pixels, enabling faster and more stable convergence; and second, Context Guidance, where we inject rich contextual priors from a large pre-trained diffusion model into the generation process. Experimental results further validate Stencil's robustness and state-of-the-art performance. We hope our work inspires future research in subject-driven generation.

\section{Acknowledgment}

This work was supported in part by the IEEE Signal Processing Society, whose financial support enabled the authors’ participation in ICIP 2025.

\bibliographystyle{IEEEbib}
\bibliography{refs}

\renewcommand{\thesubsection}{\Alph{subsection}}

\name{}
\address{}

\title{Supplementary Materials}

\maketitle  

We provide additional experimental details in Section ~\ref{sec:additional_implementation_details}. We give our user study details in Section \ref{sec:user_study}. We discuss the applications and implications of Stencil in Section \ref{sec: discussion} and provide additional qualitative results in Section \ref{sec:Additional_results}.

\subsection{Additional Implementation Details}
\label{sec:additional_implementation_details}

\subsubsection{Generating Image-Text pairs}

We use LangChain to transform GPT-4o's unstructured textual outputs into structured responses.

For each user-provided reference image, we use GPT-4o to generate a corresponding caption, forming image-text pairs which we can then use to fine-tune the U-Net backbone. We use the below system message to have GPT-4o perform the required task for us.

\begin{lstlisting}[style=grayblock, language=]

You are an professional at captioning images.
You are given some images of a subject.

You are tasked to perform the following:
1. Provide a short description of the subject, subject_name.
2. Create a detailed caption for each image containing the subject_name, image_caption. 

You are to respond in the JSON format defined below.

Format Instructions:
--------------
{format_instructions}
--------------
\end{lstlisting}

\begin{figure}[t]
    \centering
    \includegraphics[width=.99\linewidth]{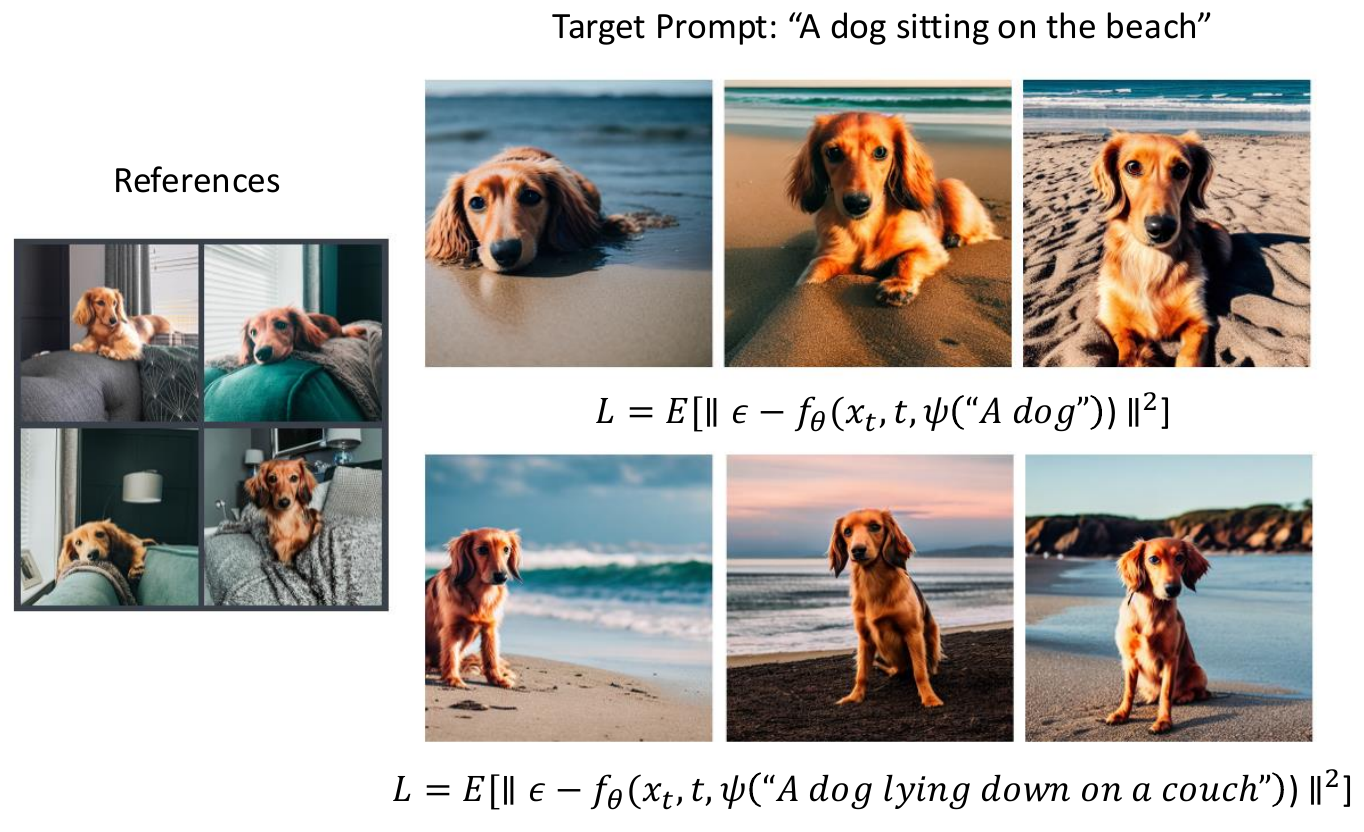}
    \captionof{figure}{\textbf{Impact of Concise vs. Detailed Prompts During Fine-Tuning.} We compare outputs from a U-Net fine-tuned on a concise caption (Top Row) versus a detailed caption (Bottom Row) of the reference images. When the caption lacks sufficient detail, the model tends to overfit to the reference image, producing less diverse generations.}

    \label{fig:action descriptors}
\end{figure}

In general, we find that fine-tuning reference images on descriptive captions yield more diverse results and is significantly less prone to overfitting compared to using concise prompts. We attribute this to language drift. When a prompt lacks sufficient detail, the model may inadvertently bind the subject tokens to both the subject’s and the background's representation. Using a more descriptive prompt helps disentangle these features, thereby improving the model’s ability to generalize. We demonstrate this point in Fig. \ref{fig:action descriptors}, where we compare the output of a U-Net fine-tuned on a set of concise captions vs a set of detailed captions.

\subsubsection{Fine-tuning the Decoder}

\begin{figure}[t]
    \centering
    \includegraphics[width=.95\linewidth]{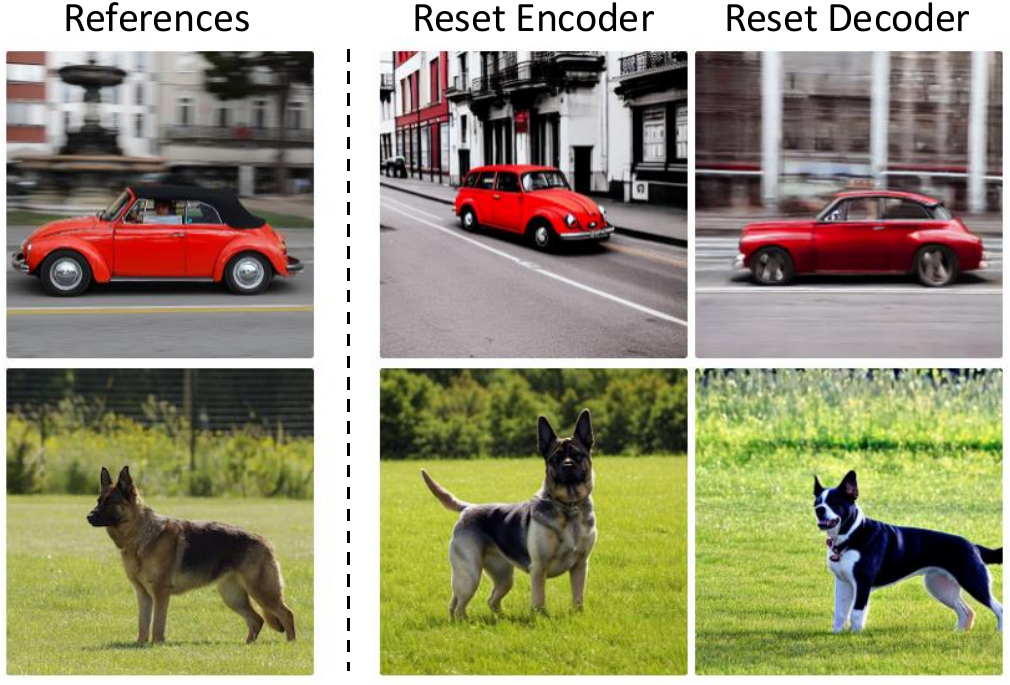}
    \captionof{figure}{\textbf{ Understanding Encoder and Decoder Learning.} We fine-tune the entire U-Net on a single reference image. Subsequently, we  reset either the encoder or the decoder by replacing their parameters with the pre-trained ones. We observed that resetting the encoder preserves the object’s appearance but causes a loss of layout, whereas resetting the decoder preserves the layout but loses fine-grain appearances.}
    \label{fig:encoder vs decoder}
\end{figure}

We demonstrate in Fig. \ref{fig:encoder vs decoder} that as spatial features propagate through the U-Net, higher-frequency information is captured. The shallower layers of the U-Net learn the structure, whereas the deeper layers learn the finer appearances of the image. Since subject-driven generation concerns the learning of higher-frequency details (e.g., appearance, color, texture, shape, etc.), we only fine-tune the U-Net decoder blocks in our implementation while freezing the rest of the network.

\begin{figure*}[htb]
            \begin{center}
                \centering
                \includegraphics[width=0.99\textwidth]{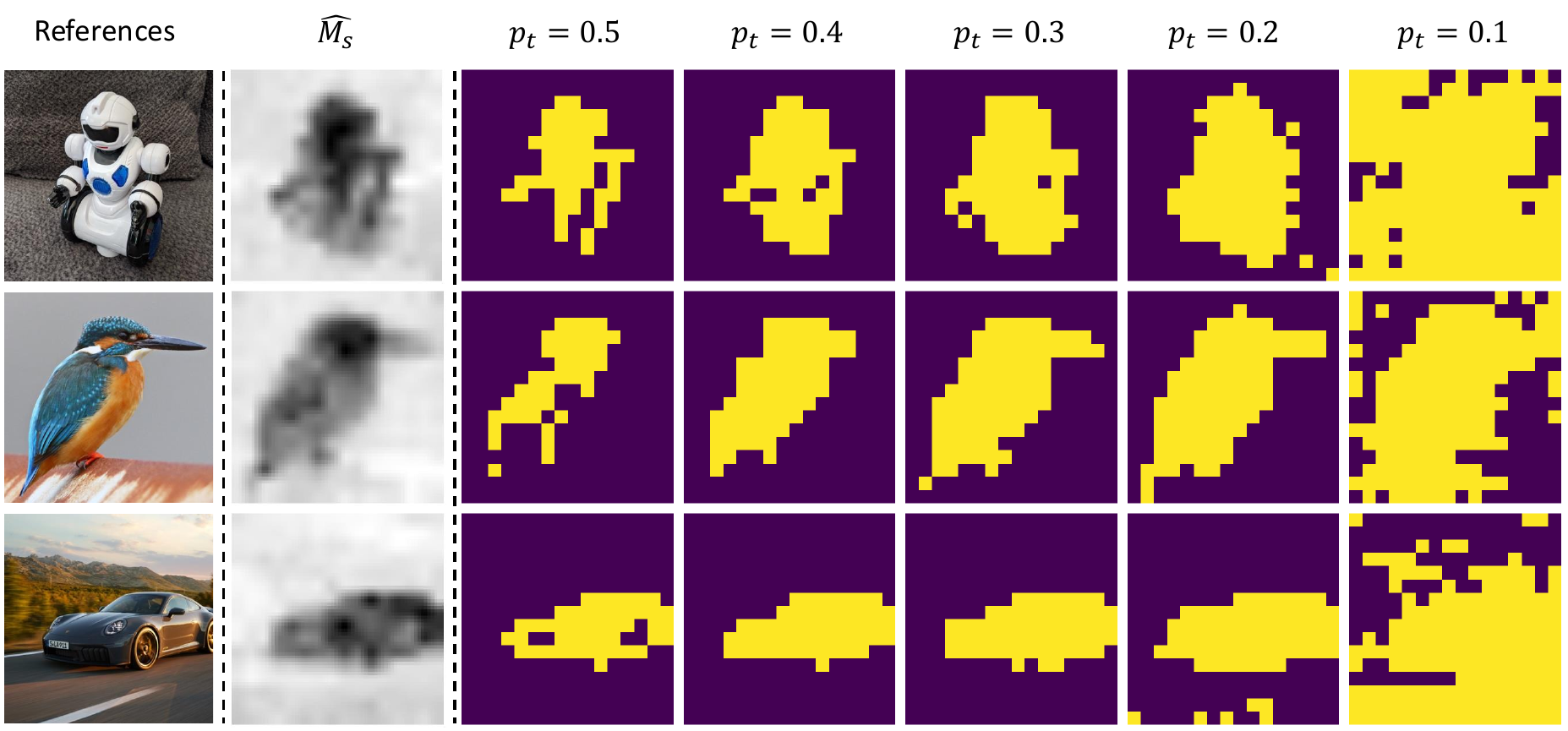}
               \captionof{figure}{\textbf{Comparing different values of $p_t$}
                }
                \label{fig:threshold}
            \end{center}
\end{figure*}

\subsubsection{Cross-Attention Guided Loss Threshold $p_t$}

We evaluate different values of the threshold $p_t$ to determine the optimal settings that maximizes the separability of subject and background pixels. This threshold represents the minimum attention weight a pixel must have toward the subject token to be considered relevant, and any pixels with an attention weight below this threshold are excluded from the loss computation. A threshold that is too low may include irrelevant background features in the loss computation, whereas a high threshold risks omitting important subject regions. Based on the results shown in Fig. \ref{fig:threshold}, we select $p_t = 0.2$.

\subsection{User Study}
\label{sec:user_study}

We conduct a user study comparing \textit{Stencil} with the previous state-of-the-art, \textit{DreamBooth}. Using the \textit{DreamBench} dataset, we evaluate all live subjects across a set of various prompts. Each image is evaluated on subject consistency and text-to-image alignment.

\vspace{0.5cm}
\begin{lstlisting}[style=grayblock, language=]
Subject Consistency: Inspect the subject of the reference image. Select which of the images best reproduces the identity of the reference subject.


Text-to-Image Alignment: Select which of the images best follows the prompt [target prompt].

If you are unsure, or believe that the images equally follow the prompt, select `Undecided'.
\end{lstlisting}

\subsection{Discussions}
\label{sec: discussion}
\subsubsection{Ethical Concerns}

 A primary concern is the potential misuse of deepfakes,which can harm reputations and spread misinformation. To mitigate these risks, greater transparency around the use and origin of AI-generated content is essential.

\subsubsection{Applications}

\begin{figure*}[htb]
            \vspace{-0.5em}
            \begin{center}
                \centering
                \vspace{-0.3in}
                \includegraphics[width=0.99\textwidth]{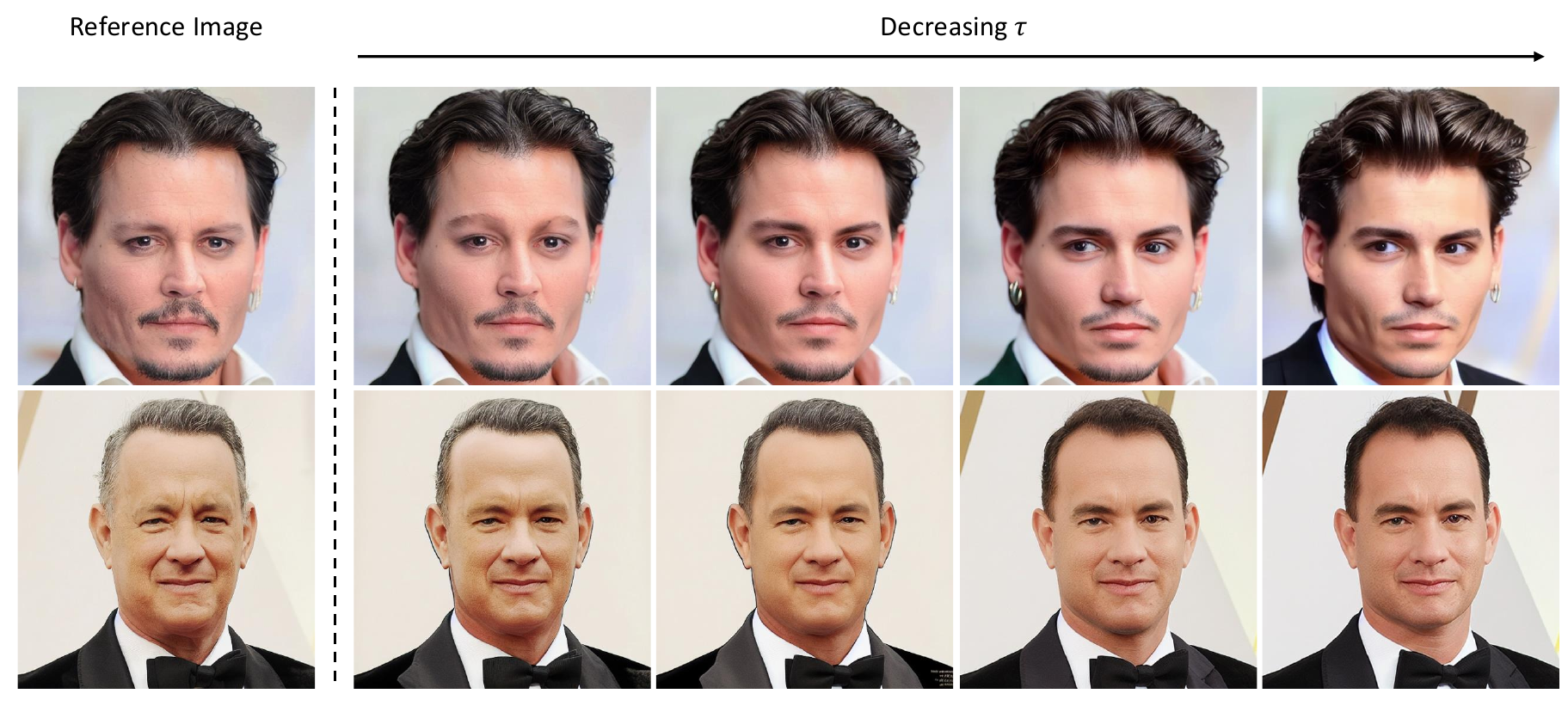}\vspace{0.1cm}
                \vskip -0.2cm
                \captionof{figure}{\textbf{Age Progression/Regression.} We can fine-tune on images of the subject’s younger self and, given a current image, interpolate observed age by adjusting the parameter $\tau$. Notably, the generated younger versions exhibit a strong resemblance to how the subjects actually appeared in their youth.
                }
                \label{Age}
            \end{center}
\end{figure*}

\begin{figure*}[htb]
            \vspace{-0.5em}
            \begin{center}
                \centering
                \vspace{-0.3in}
                \includegraphics[width=0.99\textwidth]{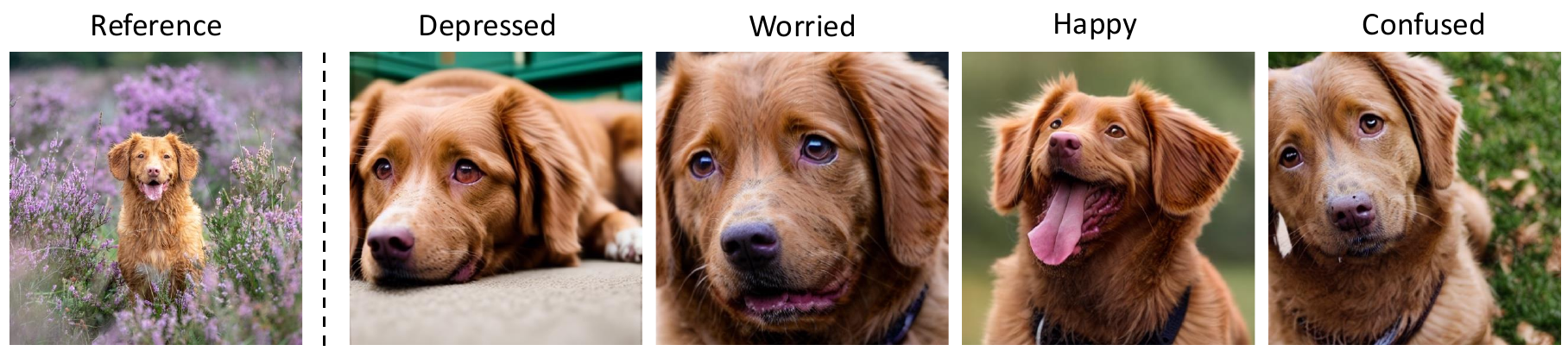}\vspace{0.1cm}
                \vskip -0.2cm
                \captionof{figure}{\textbf{Expression Manipulation.} Stencil supports the generation of a diverse range of expressions of the subject while maintaining high subject fidelity using the prompt "A [emotion] [subject token]" at inference.
                }
                \label{Expression}
            \end{center}
\end{figure*}

\begin{figure*}[htb]
            \vspace{-0.5em}
            \begin{center}
                \centering
                \vspace{-0.3in}
                \includegraphics[width=0.99\textwidth]{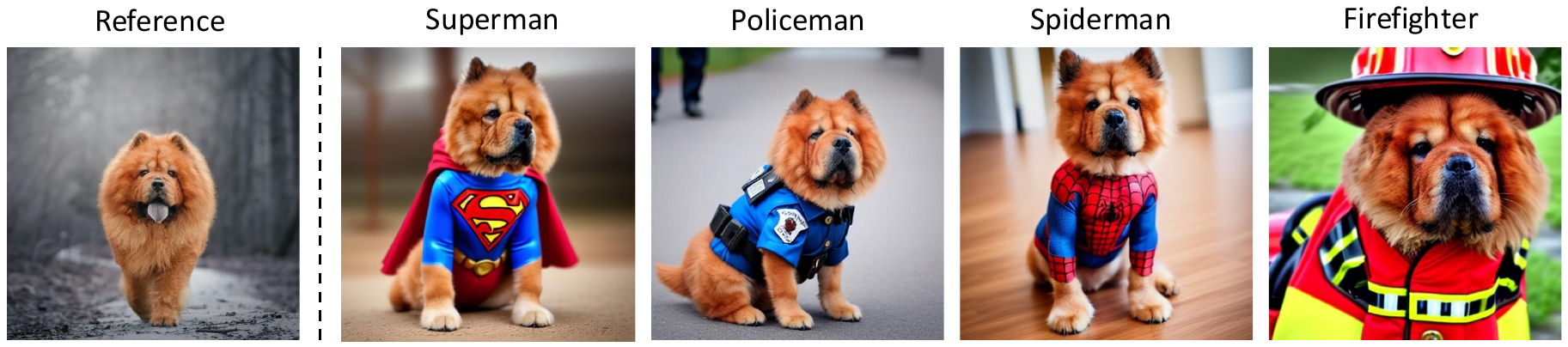}\vspace{0.1cm}
                \vskip -0.2cm
                \captionof{figure}{\textbf{Accessorization.}  We can generate the subject in various accessories by using the prompt “A [subject token] wearing [accessory].” at inference.
                }
                \label{Accessorization}
            \end{center}
\end{figure*}

\begin{figure*}[htb]
            \vspace{-0.5em}
            \begin{center}
                \centering
                \vspace{-0.3in}
                \includegraphics[width=0.99\textwidth]{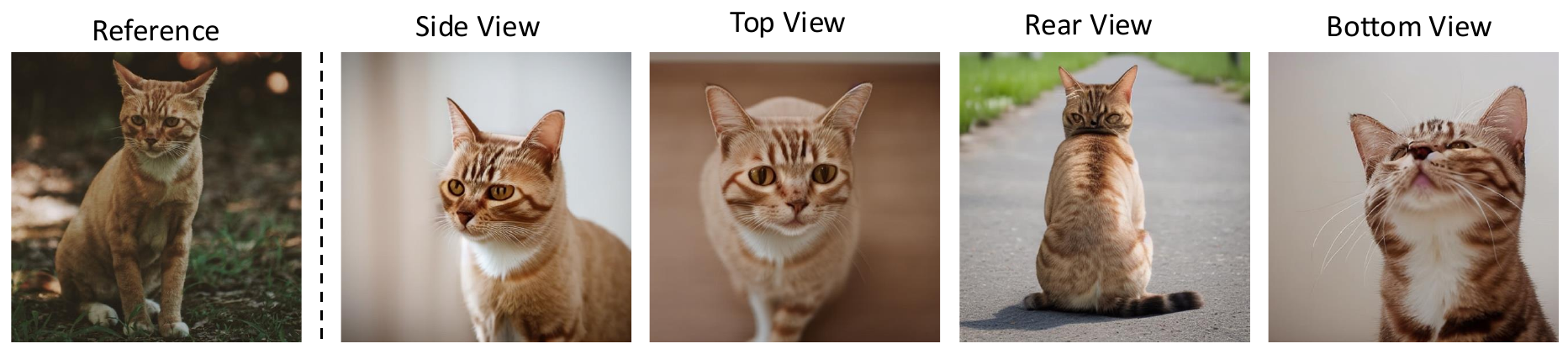}\vspace{0.1cm}
                \vskip -0.2cm
                \captionof{figure}{\textbf{Perspective-conditioned Generation.}  We can generate diverse images of the subject in different points-of-view, previously unseen in the reference images, using the prompt "A [subject token] seen from [angle]".
                }
                \label{Perspective}
            \end{center}
\end{figure*}

\begin{figure*}[htb]
            \vspace{-0.5em}
            \begin{center}
                \centering
                \vspace{-0.3in}
                \includegraphics[width=0.99\textwidth]{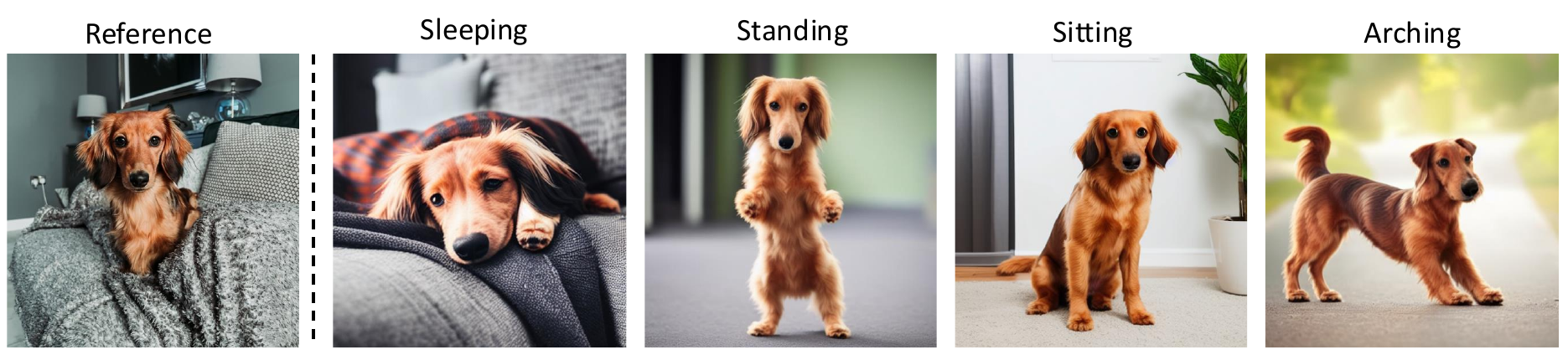}\vspace{0.1cm}
                \vskip -0.2cm
                \captionof{figure}{\textbf{Pose Editing.} Stencil can generate diverse unseen poses of the subject that is beyond the generation capabilities of the small base model. We can achieve this using the prompt “A [subject token] [pose]”. 
                }
                \label{Pose}
            \end{center}
\end{figure*}

\begin{figure*}[htb]
            \vspace{-0.5em}
            \begin{center}
                \centering
                \vspace{-0.3in}
                \includegraphics[width=0.99\textwidth]{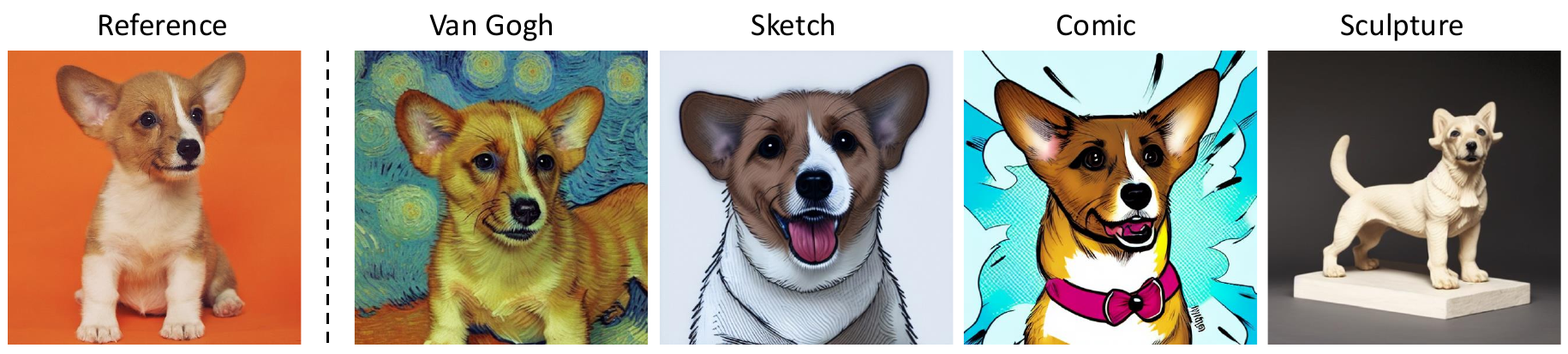}\vspace{0.1cm}
                \vskip -0.2cm
                \captionof{figure}{\textbf{Style Transfer.} Stencil enables the seamless transfer of the subject to various artistic mediums, such as paintings and sculptures while maintaining key visual characteristics using the prompt  “A [subject token] in [artistic style]”.
                }
                \label{Style transfer}
            \end{center}
\end{figure*}

Below, we present a representative (non-exhaustive) list of applications enabled by Stencil in Fig. \ref{Age}, \ref{Expression}, \ref{Accessorization}, \ref{Perspective}, \ref{Pose}, \ref{Style transfer}.

\subsection{Additional Qualitative Results}
\label{sec:Additional_results}

\begin{figure*}[htb]
            \vspace{-0.5em}
            \begin{center}
                \centering
                \vspace{-0.3in}
                \includegraphics[width=0.9\textwidth]{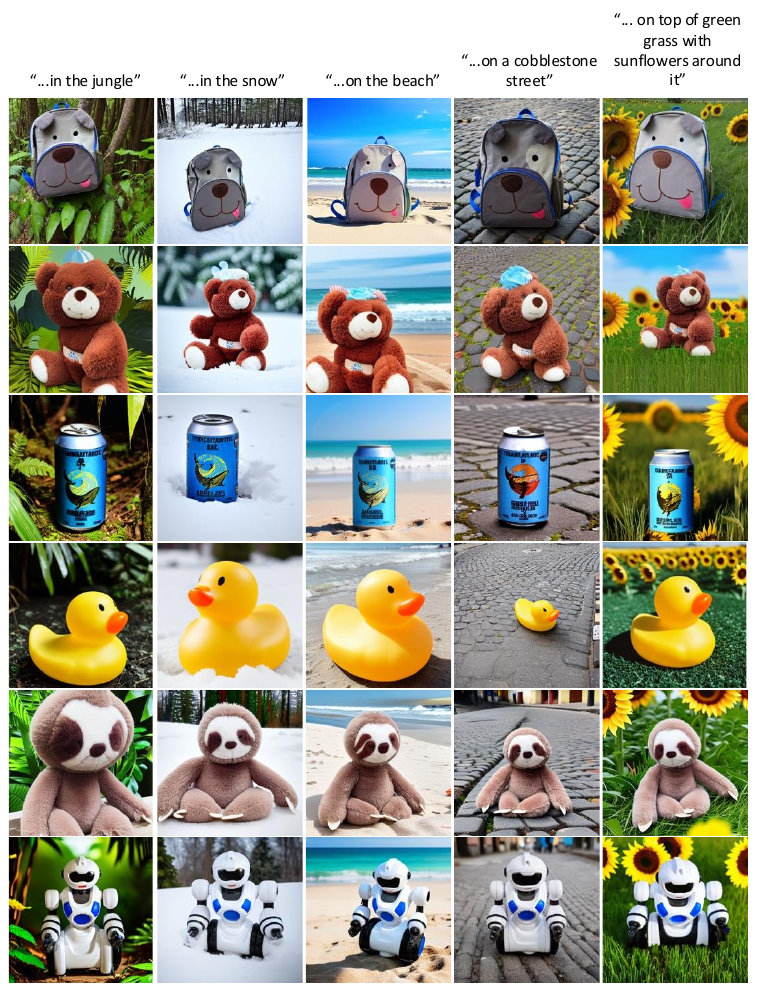}\vspace{0.1cm}
                \vskip -0.2cm
                \captionof{figure}{\textbf{DreamBench Qualitative Results Part 1}.
                }
                \label{fig:Additional_Results_1}
            \end{center}
\end{figure*}

\begin{figure*}[htb]
            \vspace{-0.5em}
            \begin{center}
                \centering
                \vspace{-0.3in}
                \includegraphics[width=0.9\textwidth]{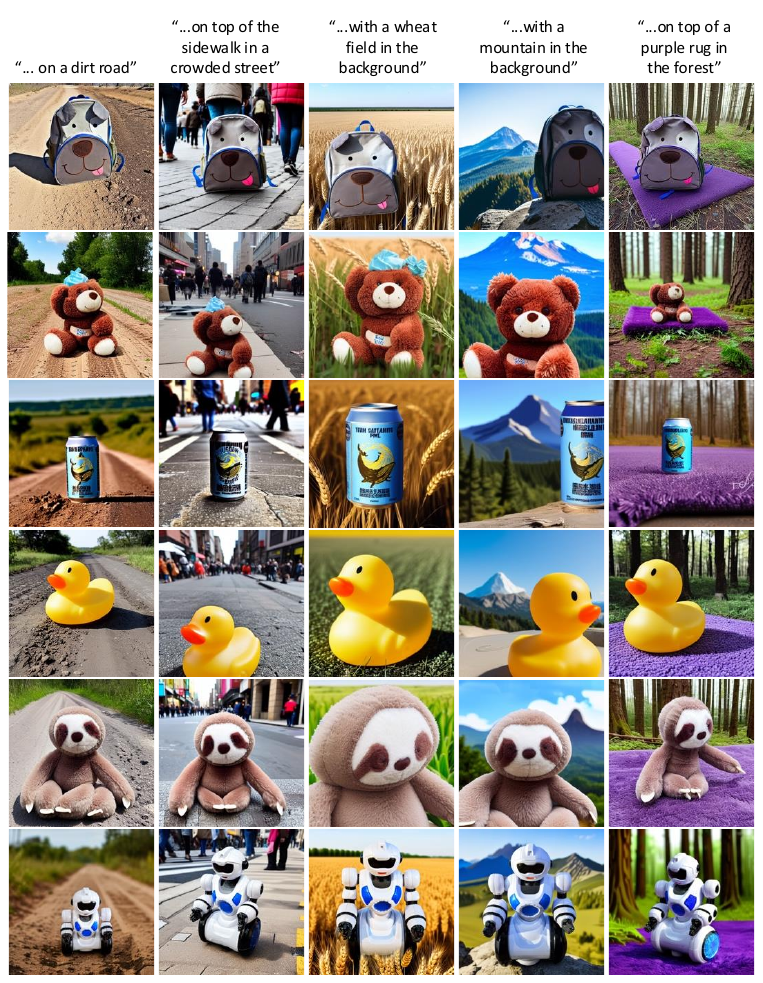}\vspace{0.1cm}
                \vskip -0.2cm
                \captionof{figure}{\textbf{DreamBench Qualitative Results Part 2}.
                }
                \label{fig:Additional_Results_2}
            \end{center}
            
\end{figure*}

We provide additional qualitative results in Fig. \ref{fig:Additional_Results_1}, \ref{fig:Additional_Results_2}.

\end{document}